\documentclass[lettersize,journal]{IEEEtran}
\usepackage{amsmath,amsfonts}
\usepackage{algorithm}
\usepackage{algpseudocode}
\usepackage{array}
\usepackage[caption=false,font=normalsize,labelfont=sf,textfont=sf]{subfig}
\usepackage{textcomp}
\usepackage{stfloats}
\usepackage{url}
\usepackage{verbatim}
\usepackage{graphicx}
\usepackage{float} 
\usepackage{booktabs} 
\usepackage{scrextend}
\usepackage{amssymb}
\usepackage{xcolor}
\usepackage{multirow}
\usepackage{amsthm}   
\usepackage{pifont}
\usepackage{wrapfig}
\definecolor{darkred}{rgb}{0.55, 0.0, 0.0}  
\definecolor{darkgreen}{rgb}{0.0, 0.5, 0.0}  
\usepackage[table]{xcolor}
\definecolor{lightgreen}{RGB}{105, 195, 105} 
\usepackage[accsupp]{axessibility}  
\usepackage{microtype}
\definecolor{Mycolor1}{HTML}{8FAADC}
\definecolor{cYellow}{HTML}{FFFFCC}
\definecolor{cRed}{HTML}{FFCCCC} 
\definecolor{cGrey}{HTML}{F3F7F2} 
\definecolor{cGreen}{HTML}{548235}
\usepackage[style=ieee,backend=bibtex]{biblatex}
\usepackage[
pdfauthor={derajan},
pdftitle={How to do this},
pdfstartview=XYZ,
bookmarks=true,
colorlinks=true,
linkcolor=blue,
urlcolor=blue,
citecolor=blue,
pdftex,
linktocpage=true,   
hyperindex=true
]{hyperref}
\makeatother
\begin{document}

\title{MoR-MLLM: Mixture of Recursions for Efficient Multimodal Large Language Models}

\author{Pengcheng Zheng, Chaoning Zhang$^{*}$, \IEEEmembership{Senior Member, IEEE}, Jiaxin Yan, Sihan Cao, Jianwei Zhang, Xudong Wang, Jiaquan Zhang, Jewon Lee, Tae-Ho Kim, Yang Yang, \IEEEmembership{Senior Member, IEEE}, and Heng Tao Shen, \IEEEmembership{Fellow, IEEE}
\thanks{This work was supported by the National Natural Science Foundation of China under grant 62572104 and 62220106008. \textit{(Corresponding author: Chaoning Zhang.)}} 
\thanks{Pengcheng Zheng, Chaoning Zhang, Jiaxin Yan, Sihan Cao, Jianwei Zhang, Jiaquan Zhang, and Yang Yang are with the Center for Future Media and School
of Computer Science and Engineering, University of Electronic Science and Technology of China, Chengdu 611731, China (e-mail: zpc777@std.uestc.edu.cn; chaoningzhang1990@gmail.com).}
\thanks{Xudong Wang is with the School of
Computing, Kyung Hee University, Yongin-si, South Korea.}
\thanks{Jewon Lee and  Tae-Ho Kim are with the Nota Inc., Seoul, South Korea.}
\thanks{Heng Tao Shen is with the School of Computer Science and Technology, Tongji University, Shanghai 200092, China.}
}

\markboth{Journal of \LaTeX\ Class Files,~Vol.~14, No.~8, August~2021}%
{Shell \MakeLowercase{\textit{et al.}}: A Sample Article Using IEEEtran.cls for IEEE Journals}


\maketitle

\begin{abstract}
  Multimodal Large Language Models (MLLMs) have demonstrated remarkable reasoning capabilities across vision and language tasks. However, their massive computational and memory demands hinder real-world deployment. While recent efforts reduce costs by employing lightweight language backbones, existing paradigms remain computation-dense due to their static sparsity and depth allocation, which cannot adapt to the semantic complexity of each token. To this end, we propose MoR-MLLM, a computation-sparse MLLM based on the recent Mixture-of-Recursions (MoR) framework. MoR-MLLM introduces adaptive per-token recursion, allowing the model to dynamically adjust its recursive depth and allocate more computation to visually or linguistically challenging tokens while skipping redundant operations for simpler ones. To stabilize the training of recursive sparsity in multimodal settings, we further design a three-stage MoR-Tuning strategy and an entropy-regularized loss to encourage diverse routing distributions. Extensive experiments show that compared with recent advanced tiny MLLMs, our proposed MoR-MLLM can greatly reduce the training memory and computation complexity while retaining high performance on various vision-language tasks. 
\end{abstract}

\begin{IEEEkeywords}
Multimodal large language models, mixture-of-recursions, token-level recursion, computation-sparse.
\end{IEEEkeywords}

\section{Introduction}
\label{sec:intro}

\IEEEPARstart{R}{ecent} advances in Multimodal Large Language Models (MLLMs), such as LLaVA~\cite{liu2023visual}, Qwen-VL~\cite{bai2023qwenvlversatilevisionlanguagemodel}, and DeepSeek-VL~\cite{lu2024deepseek}, have demonstrated remarkable abilities in unified vision-language (VL) reasoning. By coupling a vision encoder with a Large Language Model (LLM), these models can interpret complex visual scenes and perform instruction-following tasks with near-human versatility. Typically, scaling up model and dataset size has proven to be an effective strategy for performance improvement. For example, Qwen-VL-72B~\cite{bai2025qwen25vltechnicalreport} integrates a 72B-parameter language backbone with high-resolution visual encoders, while several recent works~\cite{weng2025cross, sun2026unleashing, qiang2026mitigating, zeng2025contrastive, zhang2025flexvln, zhang2023one, zhang2024unleash, zheng2026joint} extend MLLMs beyond 100B-parameters, achieving state-of-the-art (SOTA) results on a range of benchmarks. However, such scaling inevitably incurs tremendous computational and memory overhead, which limits their accessibility and deployability in real-world applications.

To reduce computational burden, recent studies have explored MLLM efficiency from complementary dimensions, including weight compression~\cite{zheng2026llava, zhang2026lightweight}, visual-token reduction~\cite{zhang2026rcp,cao2026language}, and parameter-efficient sparse adaptation~\cite{sun2026grasp, zhang2026learning}. At the architectural level, lightweight MLLMs can be broadly categorized into two paradigms: parameter-dense and parameter-sparse models.
In the parameter-dense paradigm, models improve efficiency by adopting compact architectures, yet all parameters remain active during inference. For instance, MiniCPM-V~\cite{hu2024minicpmunveilingpotentialsmall} integrates a smaller LLM with pretrained vision encoders to achieve competitive multimodal performance at a fraction of the cost. Independently, MobileVLM~\cite{wu2024mobilevlm} further introduces 1.4B and 2.7B-scale MLLMs optimized for mobile deployment. However, dense models remain inherently inefficient, since each token engages all parameters regardless of its semantic complexity. Therefore, Lin et al.~\cite{lin2024moellavamixtureexpertslarge} recently proposed the MoE-LLaVA, which is a sparse MLLM that routes each token to only a subset of experts, achieving stronger performance with fewer activated parameters.

Despite these advances, several critical issues still remain in existing methods. We argue that both current paradigms suffer from fundamental limitations that hinder improvements in efficiency and scalability. Parameter-dense models activate every weight for every token, incurring prohibitive computational and memory costs regardless of a token’s actual semantic complexity. Parameter-sparse models like MoE-LLaVA~\cite{lin2024moellavamixtureexpertslarge} alleviate this burden by conditionally activating only a subset of experts, yet their sparsity is still static: each token traverses a fixed number of layers, and the decision is made within a pre-determined depth. \textit{As a result, current MLLMs still remain computation-dense, they cannot allocate deeper processing to visually or linguistically challenging tokens while skipping redundant computation for trivial ones.}

Recently, Bae et al.~\cite{bae2025mixture} proposed the Mixture-of-Recursions (MoR) framework, which introduces a novel dimension of efficiency: \textit{adaptive token-level recursion}. Rather than distributing sparsity across multiple experts, MoR dynamically controls the recursion depth of each token, reusing a compact set of shared parameters across all recursion steps. This mechanism simultaneously achieves parameter efficiency and computation adaptivity, allowing the model to allocate more ``thinking steps” to ambiguous tokens while skipping redundant operations for easier ones. However, directly applying MoR to train computation-sparse MLLMs is non-trivial. We observe that simultaneously converting an LLM into an MLLM while introducing recursive sparsity often causes optimization instability and severe performance degradation. Through extensive analysis, we identify that the key challenge lies in initialization and training dynamics under recursive routing. To address this, we design a three-stage training framework, termed MoR-Tuning, which progressively stabilizes the transition from dense to recursive multimodal representations. Building upon this foundation, we propose MoR-MLLM, the first computation-sparse MLLM that integrates recursive parameter sharing and per-token dynamic recursion within a unified multimodal Transformer. Additionally, we introduce an entropy-regularized loss to maximize the entropy of the per-token depth distribution, thereby mitigating router collapse and promoting diverse recursion assignments.

Importantly, the methodological contribution of MoR-MLLM does not lie in recursive parameter sharing or adaptive routing individually, which are inherited from the original MoR framework. Instead, it lies in a multimodal adaptation and optimization framework for transforming a pretrained dense LLM into a computation-sparse MLLM. This transformation must jointly address visual--language alignment, initialization of the shared recursive modules from pretrained Transformer layers, and stable token-level routing. MoR-Tuning decouples these objectives through projection warm-up, joint adaptation of the grouped-shared recursive modules and routers, and end-to-end multimodal fine-tuning. Together with entropy-regularized routing, this progressive strategy preserves multimodal alignment while enabling token-adaptive recursive computation.

To conclude, the main contributions of this paper are summarized as follows:

\begin{enumerate}
\item[$\bullet$] We present MoR-MLLM, the first MoR-based computation-sparse MLLM, which integrates recursive parameter sharing and adaptive per-token recursion routing mechanisms for efficient MLLMs.

\item[$\bullet$] We propose MoR-Tuning, a novel three-stage progressive training strategy that enables stable adaptation of recursive sparsity to multimodal settings and mitigates optimization degradation caused by naive sparsification.

\item[$\bullet$] An entropy-regularized loss is introduced to maximize the routing distribution’s entropy and prevents degenerate ``single-depth'' behavior, ensuring balanced computation allocation across tokens.

\item[$\bullet$] Extensive experiments across 9 multimodal benchmarks demonstrate that MoR-MLLM achieves SoTA trade-offs between performance and efficiency, reducing training memory and inference FLOPs by over 50\% while maintaining competitive accuracy compared with recent lightweight MLLMs.
\end{enumerate}

\section{Related Work}
\label{sec:formatting}
\subsection{Recursive Models}
Recursive models have long been explored for improving parameter efficiency and memory optimization in deep neural networks. Dehghani et al.~\cite{dehghani2018universal} have pioneered the Universal Transformer, which employs a self-attention–based recurrent mechanism to reduce parameter redundancy. Following this, Looped Transformers~\cite{yang2023looped} optimize recursive loops to better simulate iterative reasoning algorithms. However, both methods adopt a fixed recursion depth, resulting in inefficient computation and insufficient expressiveness for semantically complex tokens. To alleviate these limitations, Fixed Relaxed Recursive Transformers~\cite{bae2024relaxed} introduce LoRA adapters~\cite{hu2022lora} to relax the recursion constraints within the key–value cache, enhancing flexibility in both computation and memory usage. Nevertheless, the recursion depth in these models remains static and cannot adapt dynamically to token complexity. Recently, Bae et al.~\cite{bae2025mixture} proposed the Mixture-of-Recursions (MoR) framework, which introduces lightweight routing modules to achieve dynamic, per-token recursive depth. MoR combines efficient key–value caching with unified parameter sharing, significantly improving both computational and storage efficiency compared to prior recursive designs. Concurrently, RecursiveVLM~\cite{xu2026looping} also extends recursive Transformers to multimodal understanding, yet it adopts a fixed recursion depth coupled with a monotonic recursion loss, rather than token-adaptive routing.

In this work, we are the first to leverage Mixture of Recursions for making more efficient MLLMs.

\subsection{Mulimodal Large Language Models}
Mulimodal Large Language Models (MLLMs) have recently shown remarkable performance in unified visual–linguistic reasoning tasks. LLaVA~\cite{liu2023visual} pioneers visual instruction tuning, combining a vision encoder and an LLM for general-purpose multimodal understanding. Qwen-VL~\cite{bai2023qwenvlversatilevisionlanguagemodel} further advances this paradigm by improving visual–text alignment through large-scale training and cross-domain localization tasks. Although Qwen-VL enhances image resolution and versatility, its relatively compact vision backbone limits deeper feature extraction. To improve generalization and scalability, InternVL~\cite{chen2024internvl} scales the vision encoder to 6B-parameters and aligns it with large-scale web data, achieving strong cross-domain performance. However, it lacks optimization for computational efficiency under real-world dynamic scenarios. Subsequently, DeepSeek-VL~\cite{lu2024deepseek} proposes a balanced training strategy for high-resolution images and textual reasoning, achieving efficient visual–linguistic processing in diverse conditions. 

Departing from existing approaches, MoE-LLaVA~\cite{lin2024moellavamixtureexpertslarge} is the first to incorporate a Mixture-of-Experts routing mechanism into multimodal LLMs, enabling token-wise conditional computation that significantly improves efficiency without sacrificing accuracy. Unlike MoE-LLaVA~\cite{lin2024moellavamixtureexpertslarge}, our MoR-MLLM performs adaptive per-token recursion, allocating computation depth based on token complexity for finer-grained efficiency.
\begin{figure*}[!t]
    \centering
    \includegraphics[width=1\linewidth]{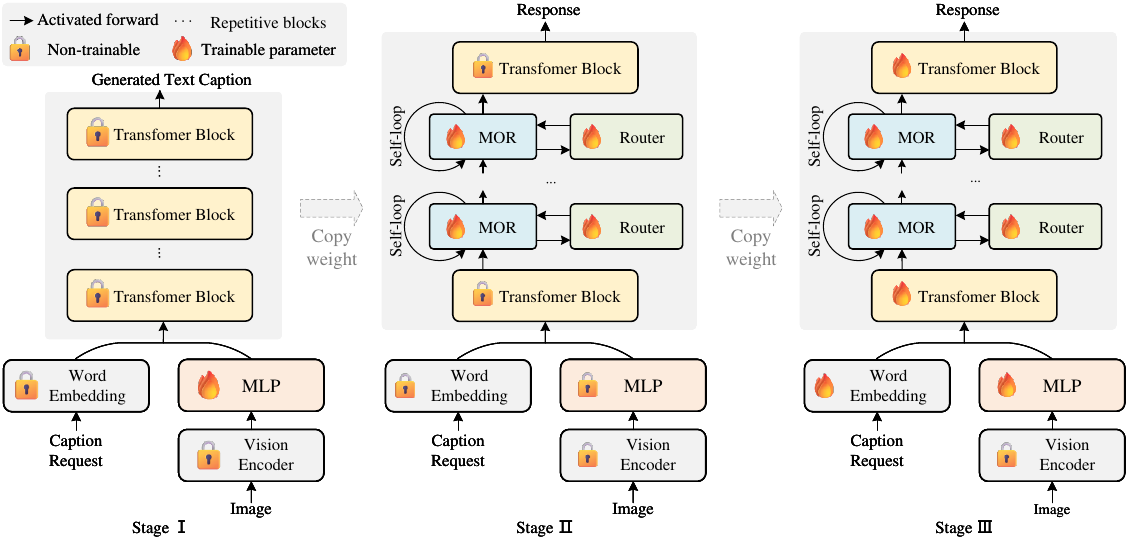}
\caption{Training framework and strategy. MoR-MLLM adopts a three-stage protocol. Stage I: Projection Warm-up: During this stage, only the MLP parameters are optimized to enable the LLM to process visual inputs.
Stage II: Router warm-up: We replace selected LLM layers with initialized MoR blocks and train only the MoR shared submodules \(\mathcal{M}(\cdot) \) and their routers. 
Stage III: Joint fine-tuning: We unfreeze all the Transformer blocks, word embedding module and MLP. Then we fine-tune them
jointly with the router on downstream multimodal tasks.}
    \label{fig:MoR-MLLM}
\end{figure*}
\begin{figure}[!t]
 \includegraphics[width=\linewidth]{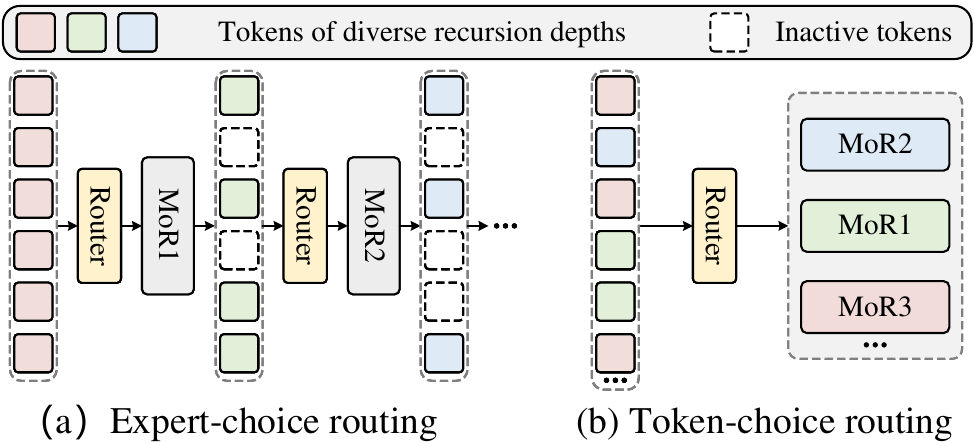}
\caption{Overview of MoR routing. At each recursion round, tokens not selected by the router are removed from further recursion and forwarded to the next downstream layer.}
\label{ExpertChoicevsTokenChoice}
\end{figure}
\section{Method}
\label{sec:method}
As illustrated in Fig.~\ref{fig:MoR-MLLM}, MoR-MLLM comprises a vision encoder, an MLP-based visual projection layer, a word embedding layer, a stack of LLM blocks, and multiple MoR blocks. We first provide preliminary knowledge on MoR in Section~\ref{subsec:mor_concise}. Then, Section~\ref{sec:architecture_new} describes the overall architecture of MoR-MLLM, including the design of the MoR block and its routing strategy. Section~\ref{sec:training} presents the three-stage MoR-Tuning strategy for stable training and progressive sparsity adaptation. Finally, in Section~\ref{sec:loss}, we elaborate on the loss functions of MoR-MLLM.
\subsection{Preliminary knowledge on MoR}
\label{subsec:mor_concise}
In each MoR block, a compact set of shared parameters is reused across multiple recursive steps. During both pre-training and inference, MoR dynamically determines the recursion depth of each token through a lightweight routing mechanism. As illustrated in Fig.~\ref{ExpertChoicevsTokenChoice}, two routing paradigms are available: expert-choice routing and token-choice routing.

\noindent\textbf{Expert-choice routing.}  
Let \(z_{t}^{(s)}\in\mathbb{R}^{D}\) denote the hidden representation of token \(t\) at recursion step \(s\). For each step \(s\), a step-specific router \(\rho_{s}(\cdot)\) produces a scalar routing score:
\begin{equation}\label{eq:expert_score_new}
\alpha_{t}^{(s)} \;=\; \Phi\bigl(\rho_{s}(z_{t}^{(s)})\bigr),
\end{equation}
where \(\Phi:\mathbb{R}\to(0,1)\) is a bounded activation (e.g. sigmoid). Denote the collection of scores at step \(s\) by \(\mathcal{A}^{(s)}=\{\alpha_{1}^{(s)},\dots,\alpha_{T}^{(s)}\}\) and let the adaptive gating threshold be the \(\gamma\)-quantile of \(\mathcal{A}^{(s)}\):
\begin{equation}\label{eq:quantile_new}
\tau^{(s)} \;=\; \operatorname{Quantile}_{\gamma}\bigl(\mathcal{A}^{(s)}\bigr).
\end{equation}
Only tokens with \(\alpha_{t}^{(s)}>\tau^{(s)}\) are forwarded to the shared recursive sub-module \(\mathcal{M}(\cdot;\Theta)\). Therefore, the update of the hidden-state is formulated as:
\begin{equation}\label{eq:expert_update_new}
z_{t}^{(s+1)} \;=\;
\begin{cases}
\;\alpha_{t}^{(s)}\,\mathcal{M}\bigl(z_{t}^{(s)};\Theta\bigr)\;+\; z_{t}^{(s)}, & \text{if }\; \alpha_{t}^{(s)}>\tau^{(s)},\\[6pt]
\;z_{t}^{(s)}, & \text{otherwise},
\end{cases}
\end{equation}
where \(\mathcal{M}(\cdot;\Theta)\) is the shared Transformer sub-module inside the MoR block and \(\Theta\) is its parameters. This mechanism concentrates computation on tokens judged to benefit most from additional recursion.


\vspace{0.5em}
\noindent\textbf{Token-choice routing.}  
Under token-choice routing, each token determines its target recursion depth at initialization. From the token's initial representation \(z_{t}^{(1)}\), a set of per-depth routers \(\{\psi_{j}(\cdot)\}_{j=1}^{M}\) produce depth-specific scores:
\begin{equation}\label{eq:token_logits_new}
\beta_{t}^{(j)} \;=\; \Phi\bigl(\psi_{j}(z_{t}^{(1)})\bigr),\qquad j\in\{1,\dots,M\},
\end{equation}
and the token selects its depth by:
\begin{equation}\label{eq:depth_selection_new}
d_{t} \;=\; \arg\max_{1\le j\le M}\, \beta_{t}^{(j)}.
\end{equation}
Token \(t\) is then recursively processed for exactly \(d_{t}\) steps. Let \(\alpha_{t}^{(s)}\) denote the stepwise intensity, which is directly derived from \(\{\beta_{t}^{(j)}\}\). The hidden-state recursion for \(s=1,\dots,d_{t}\) is written as:
\begin{equation}\label{eq:token_update_new}
z_{t}^{(s+1)} \;=\;
\begin{cases}
\;\alpha_{t}^{(s)}\,\mathcal{M}\bigl(z_{t}^{(s)};\Theta\bigr)\;+\; z_{t}^{(1)}, & \text{if } s = d_{t},\\[6pt]
\;\alpha_{t}^{(s)}\,\mathcal{M}\bigl(z_{t}^{(s)};\Theta\bigr), & \text{if } 1\le s < d_{t}.
\end{cases}
\end{equation}
The residual connection to the initial embedding \(z_{t}^{(1)}\) at the final step helps preserve the original token content while allowing depth-dependent refinement.
Both paradigms reuse the same recursive sub-module \(\mathcal{M}(\cdot;\Theta)\) but differ in when and how a token's recursion budget is determined. Expert-choice routing decides continuation at each step (adaptive, stepwise selection), whereas token-choice routing commits to a depth at the outset (single-shot assignment).


\subsection{MoR-MLLM Architecture and Adaptive Recursion}
\label{sec:architecture_new}

\begin{figure}[!t]
 \includegraphics[width=\linewidth]{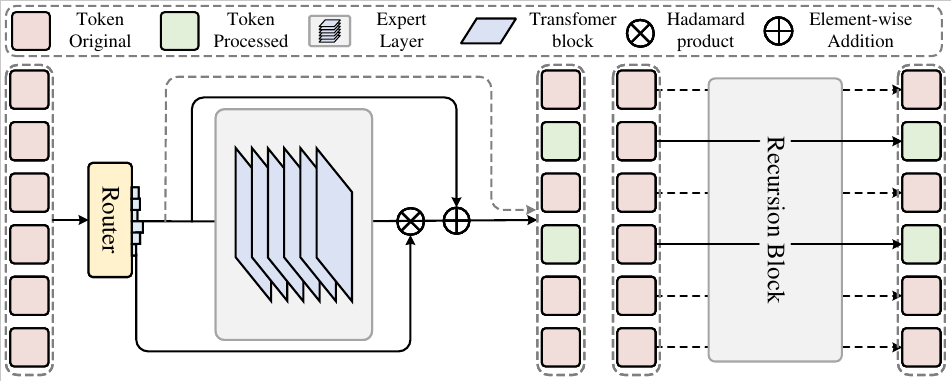}
 \caption{Illustration of the MoR block and routing mechanism. Each router computes token-level scores to decide whether a token proceeds to the next recursion step. Selected tokens are processed recursively by the shared Transformer, while unselected tokens skip further updates.}
\label{MoRblock}
\end{figure}

Our proposed MoR-MLLM integrates MoR blocks into a standard MLLM to achieve adaptive computation. As shown in Fig.~\ref{fig:MoR-MLLM} (Stage I), the base MLLM consists of a vision encoder, a visual projection layer, a word embedding layer, and multiple Transformer layers.
Given a RGB image $v\in\mathbb{R}^{H\times W\times 3}$, the visual encoder produces a sequence of visual tokens $Z=[z_1,\dots,z_P]\in\mathbb{R}^{P\times C}$, where $P$ denotes the number of visual patches. A visual projection MLP $f_{\text{proj}}(\cdot)$ maps $Z$ to the LLM hidden dimension:
\begin{equation}\label{eq:Vproj}
V = f_{\text{proj}}(Z) \in \mathbb{R}^{P\times D},
\end{equation}
where $D$ is the hidden dimension of the LLM. The text input is embedded by a word embedding layer $g(\cdot)$ to produce text embeddings $T=[t_1,\dots,t_N]\in\mathbb{R}^{N\times D}$, where $N$ is the text sequence length. We then concatenate the visual and textual tokens along the sequence dimension and use the resulting sequence as the input to the subsequent blocks:
\begin{equation}\label{eq:x0}
X = [v_1,\dots,v_P,t_1,\dots,t_N] \in \mathbb{R}^{(P+N)\times D},
\end{equation}
where \(v_i=f_{\text{proj}}(z_i)\in\mathbb{R}^D\) and \(t_j=g(w_j)\in\mathbb{R}^D\), $X$ represents the input token sequence to the subsequent Transformer blocks. As shown in Fig.~\ref{fig:MoR-MLLM} (Stage II), we systematically replace blocks of intermediate Transformer layers with our proposed MoR blocks. Specifically, for an LLM with a total of \( n \) Transformer layers, we preserve the first and the last layer intact. The remaining intermediate \( n-2 \) layers are evenly divided into \( m \) groups, and each group is replaced by a single MoR block. This results in a total of \( k \) Transformer layers being integrated into an MoR block, where the number of blocks is given by $k = \frac{n - 2}{m}$. Each MoR block comprises a shared Transformer module $\mathcal{M}(\cdot)$ and a lightweight router $r(\cdot)$ that determines per-token recursion behavior.
\noindent\textbf{MoR block forward.}
As illustrated in Fig.~\ref{ExpertChoicevsTokenChoice}, the MoR framework supports two routing mechanisms: \textit{token choice} and \textit{expert choice} routing. We adopt the \textit{expert-choice} variant, which has been shown to yield superior performance as it stabilizes optimization by decoupling routing decisions across recursion steps and allows token-wise adaptive depth selection without early commitment~\cite{bae2025mixture}. As shown in Fig.~\ref{MoRblock}, the MoR block incorporates adaptive recursive computation, where a shared Transformer sub-module \( \mathcal{M}(\cdot) \) is selectively applied to tokens according to routing decisions. At each recursion step \(s\), a lightweight router network \(r(\cdot)\) computes a scalar routing signal \(\lambda_t^{(s)}\) for each token \(t\) from its current hidden state \(\mathcal{H}_t^{(s)}\). The routing signal \(\lambda_t^{(s)}\) determines whether token \(t\) will be processed by the shared Transformer sub-module \(\mathcal{M}(\cdot)\) at the next recursion. If a token is not selected, its hidden state remains unchanged. This mechanism enables adaptive allocation of computation to tokens that benefit most from deeper processing. The forward process is formulated as:
\begin{equation}\label{eq:new_router_score}
\lambda_t^{(s)} \;=\; \Psi\!\bigl(r(\mathcal{H}_t^{(s)})\bigr),
\end{equation}
\begin{equation}\label{eq:new_recursion_update}
\mathcal{H}_t^{(s+1)} \;=\;
\begin{cases}
\;\lambda_t^{(s)} \cdot \mathcal{M}\bigl(\mathcal{H}_t^{(s)}\bigr) + \mathcal{H}_t^{(s)}, & \text{if } \lambda_t^{(s)} > \mathcal{Q}_{\beta}\!\bigl(\Lambda^{(s)}\bigr),\\[6pt]
\;\mathcal{H}_t^{(s)}, & \text{otherwise},
\end{cases}
\end{equation}
where \(\Lambda^{(s)}=\{\lambda_1^{(s)},\dots,\lambda_{T}^{(s)}\}\) is the multiset of routing signals at step \(s\), \(\Psi(\cdot)\) denotes a bounded activation (i.e., sigmoid) that maps router outputs to a gating interval, and \(\mathcal{Q}_{\beta}(\Lambda^{(s)})\) returns the \(\beta\)-th percentile of the set \(\Lambda^{(s)}\) (ablated in Section~\ref{thre}). The recursion proceeds for up to $S$ steps, where $\mathcal{H}_t^{(S)}$ denotes the final token representation produced by the MoR block for each token.

\subsection{Three-Stage Recursive Adaptation}\label{sec:training} 

To stabilize optimization and progressively introduce recursive capacity, we adopt a three-stage training protocol. The overall process is illustrated in Fig.~\ref{fig:MoR-MLLM}, where each stage addresses a specific challenge in adapting MoR blocks to multimodal learning. 
The central role of MoR-Tuning is to decouple three objectives that are entangled in a naive multimodal adaptation of MoR: visual--language alignment, adaptation of the shared recursive modules and routers, and downstream multimodal specialization. The three stages therefore introduce recursive sparsity progressively rather than optimizing all components jointly from the beginning.

\noindent\textbf{Stage I: Projection Warm-up.}  
The primary challenge at this stage is to align visual features with the input space of the LLM. Specifically, a lightweight MLP maps the image tokens into the LLM embedding space, where the image patches are represented as pseudo-text tokens. This step allows the LLM to interpret visual inputs without modifying its original parameters. At this stage, MoR blocks are not yet applied.
\noindent\textbf{Stage II: Router Warm-up.} Once image tokens are aligned, we introduce MoR blocks by replacing selected LLM layers. To ensure stability, $\mathcal{M}(\cdot)$ is initialized using the grouped sharing strategy~\cite{ravanbakhsh2017equivariance}, which is ablated in Section~\ref{ablation}. During this stage, both the shared Transformer module $\mathcal{M}(\cdot)$ and the lightweight router are jointly trained to adapt the recursive computation mechanism and learn reasonable recursion probabilities. A negative bias is used for the router to encourage shallow recursion early in training.

\noindent\textbf{Stage III: Joint Fine-tuning.}  
After the router has stabilized, we unfreeze the shared Transformer module \(\mathcal{M}(\cdot) \) and fine-tune it jointly with the router on downstream multimodal tasks. Parameter-efficient adapters, i.e., LoRA adapter~\cite{hu2022lora} is combined to improve performance. At this stage, additional regularization terms on expected recursion depth, i.e, the entropy-regularization is also introduced to encourage more diverse routing distributions.

\subsection{Loss Functions}\label{sec:loss}
Since the introduction of MoR blocks brings architectural modifications to the original MLLM, the model may struggle to fully adapt to these changes without additional guidance. To mitigate this, we design a training objective that integrates task-specific supervision with a regularization term encouraging diverse routing behaviors. The overall objective is defined as:
\begin{equation}\label{eq:total_loss}
    \mathcal{L} = \mathcal{L}_{\mathrm{task}} + \lambda_{\mathrm{ent}} \cdot \mathcal{L}_{\mathrm{ent}}.
\end{equation}
Here, $\lambda_{\mathrm{ent}}$ controls the strength of the regularization term.

\noindent\textbf{Task loss.}  
$\mathcal{L}_{\mathrm{task}}$ denotes the downstream supervision signal, such as cross-entropy for language modeling or other multimodal training objectives.

\noindent\textbf{Entropy regularization.}  
To prevent degenerate routing distributions where all tokens are assigned to the same depth, we encourage lower negative entropy, i.e., maximize the entropy, in the router's output distribution:
\begin{equation}\label{eq:entropy}
    \mathcal{L}_{\mathrm{ent}} = \frac{1}{T}\sum_{i=1}^T \sum_{k=1}^{S} \pi(k \mid x_i) \log \pi(k \mid x_i),
\end{equation}
where $T$ is the total number of tokens, $S$ is the maximum recursion depth (ablated in Section~\ref{depth}), and $\pi(k\mid x_i)$ is the router probability of assigning token $x_i$ to depth $k$. This term promotes diversity in depth assignments, which improves training stability and model robustness.

\section{Experiments}
\label{sec:exp}
\subsection{Training Configuration and Model Setup}

Following~\cite{lin2024moellavamixtureexpertslarge}, we adopt the same VL backbone, visual projector, input resolution, and stage-wise data organization, while replacing the MoE layers with the proposed MoR blocks and independently configuring the optimization schedule for recursive routing. Specifically, we adopt
CLIP-Large/336~\cite{radford2021learning} as the vision
encoder and Phi-2 (2.7B)~\cite{phi2} as the language
backbone. The image projector consists of two
MLP~\cite{rumelhart1986learning} layers with GeLU
activation~\cite{hendrycks2016gaussian}. The first and last
Transformer layers are retained as standard Transformer
layers, while the 30 intermediate layers are evenly
partitioned into $m$ groups, each replaced by a MoR block comprising a shared Transformer sub-module $\mathcal{M}(\cdot)$ and a lightweight router $r(\cdot)$ implemented as a linear projection with sigmoid activation. Based on ablation results in Section~\ref{ablation}, $m$ is set to 5, each corresponding to
\(k=6\) original Transformer layers. The routing threshold $P_{\beta}$ and maximum recursion depth $s$ are set to 80\% and 32, respectively, and the balancing coefficient $\lambda_{\mathrm{ent}}$ is fixed at 0.10.

We optimize MoR-MLLM using AdamW~\cite{loshchilov2017decoupled}
with no weight decay and a cosine learning-rate
schedule~\cite{cosin}. Stage I is trained for 10K steps
using a learning rate of \(1\times10^{-4}\), a global batch
size of 64, and a warm-up ratio of 0.03. Stage II is trained
for 20K steps using a learning rate of \(5\times10^{-5}\),
a global batch size of 64, and a warm-up ratio of 0.05.
Stage III is trained for 30K steps using a learning rate of
\(2\times10^{-5}\), a global batch size of 32, and a warm-up
ratio of 0.10. Across all three stages, the input images are
resized to \(336\times336\), the maximum text sequence
length is set to 2048, and Bf16 mixed precision is employed.
Training is conducted on eight NVIDIA RTX 4090 GPUs. We use DeepSpeed
ZeRO-2~\cite{hanindhito2024bandwidth} in Stages I and II
and enable ZeRO-2 offloading in Stage III to accommodate
the increased memory requirements of joint fine-tuning.
Following the stage-wise data composition adopted by
MoE-LLaVA~\cite{lin2024moellavamixtureexpertslarge}, Stage I
performs projection warm-up using the 558K-sample
LLaVA-Pretrain corpus~\cite{blip} to establish initial
VL alignment. Stage II performs router warm-up
on a 964K-sample hybrid corpus comprising 157K samples
from SViT~\cite{zhao2023svit}, 220K from
LVIS~\cite{wang2023see}, 331K from
LRV~\cite{liu2023aligning}, and 256K from
MIMIC-IT~\cite{li2023mimic}. Stage III jointly fine-tunes
the multimodal backbone, MoR blocks, and routers on the
665K-sample LLaVA-Mix665k instruction-tuning
dataset~\cite{liu2024improved}. This organization preserves
the established multimodal training pipeline for controlled
comparison while progressively introducing and optimizing
the proposed recursive computation mechanism.
\begin{figure*}[!t]
    \centering
    \includegraphics[width=1\linewidth]{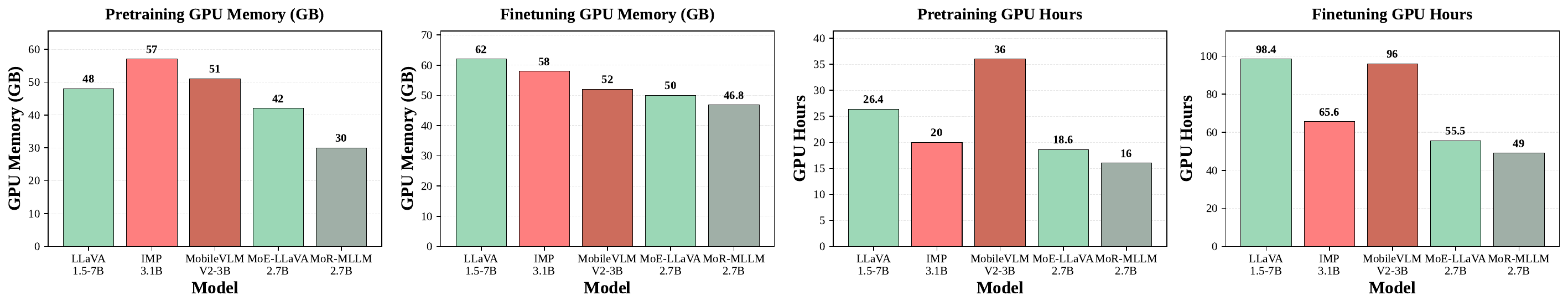}
\caption{Comparison of training efficiency between MoR-MLLM and recent lightweight MLLMs. Following~\cite{tong2025flashsloth},
we evaluate the training efficiency of MoR-MLLM on
eight NVIDIA RTX 4090 GPUs and report GPU memory usage
and GPU-hour cost during pretraining and fine-tuning. MoR-MLLM requires less GPU memory and fewer GPU hours during both stages, demonstrating its
superior training efficiency. The
LLaVA pretraining and LLaVA-665k instruction-tuning
splits~\cite{liu2024improved} are used for fair comparisons.
}
  \label{training}
\end{figure*}

\begin{table*}[!t]
\centering
\caption{
Comparison of inference efficiency between MoR-MLLM and recent
lightweight MLLMs across five representative tasks.
Following FlashSloth~\cite{tong2025flashsloth}, we evaluate the
inference efficiency of MoR-MLLM using five metrics: the number
of visual tokens, first-round inference TFLOPs, GPU memory usage
(GB), first-token response latency (seconds), and throughput
(samples/second). The best results are \textbf{bold}, and the
second-best results are \underline{underlined}.
}
\label{inference effeciency}

\footnotesize
\setlength{\tabcolsep}{2.0pt}
\renewcommand{\arraystretch}{1.10}

\resizebox{\textwidth}{!}{
\begin{tabular}{lccccc|ccccc|ccccc}
\toprule

\multicolumn{1}{c}{} &
\multicolumn{15}{c}{
\textit{Inference efficiency across multimodal benchmarks}
} \\
\cmidrule(lr){2-16}

\multirow{2}{*}{\textbf{Model}} &
\multicolumn{5}{c|}{\textbf{MME}} &
\multicolumn{5}{c|}{\textbf{MMB}} &
\multicolumn{5}{c}{\textbf{POPE}} \\
\cmidrule(lr){2-6}
\cmidrule(lr){7-11}
\cmidrule(lr){12-16}

&
\textbf{\#Tok.} &
\textbf{TFLOPs} &
\textbf{Lat.} &
\textbf{Mem.} &
\textbf{Thpt.} &
\textbf{\#Tok.} &
\textbf{TFLOPs} &
\textbf{Lat.} &
\textbf{Mem.} &
\textbf{Thpt.} &
\textbf{\#Tok.} &
\textbf{TFLOPs} &
\textbf{Lat.} &
\textbf{Mem.} &
\textbf{Thpt.} \\
\midrule

LLaVA V1.5-7B~\cite{liu2024improved}
& \underline{576} & 4.43 & \underline{0.11} & 15.3 & 5.6
& 576 & 4.77 & 0.11 & 16.4 & 6.1
& 576 & 4.40 & \underline{0.11} & 15.3 & 7.8 \\

IMP-3.1B~\cite{shao2025imp}
& 729 & \underline{2.40} & 0.12 & 8.4 & \underline{9.1}
& 729 & 2.40 & 0.12 & 8.4 & \underline{8.4}
& 729 & 2.40 & 0.12 & 8.4 & \underline{9.4} \\

Qwen2-VL-2B~\cite{wang2024qwen2}
& 646 & 2.72 & 0.41 & 59.2 & 1.8
& \textbf{385} & \underline{0.84} & \underline{0.09}
& \underline{8.0} & 7.0
& \textbf{353} & \underline{1.51} & 0.13 & 7.8 & 5.0 \\

InternVL2-2B~\cite{internvl}
& 1478 & 4.45 & 0.21 & 10.0 & 3.7
& 1296 & 3.98 & 0.19 & 11.5 & 4.7
& 1666 & 4.99 & 0.23 & \underline{7.4} & 3.8 \\

FlashSloth-HD-3.2B~\cite{tong2025flashsloth}
& \textbf{414} & 2.72 & \underline{0.11}
& \underline{7.8} & 7.3
& \underline{414} & 1.98 & 0.11 & 8.7 & 5.1
& \underline{414} & 2.71 & \underline{0.11} & 7.8 & 7.3 \\

\textbf{MoR-MLLM-2.7B (Ours)}
& \underline{\textbf{576}} & \textbf{0.68}
& \textbf{0.05} & \textbf{7.1} & \textbf{15.6}
& 576 & \textbf{0.52} & \textbf{0.04}
& \textbf{7.0} & \textbf{17.2}
& 576 & \textbf{0.45} & \textbf{0.05}
& \textbf{7.2} & \textbf{18.5} \\

\midrule

\multirow{2}{*}{\textbf{Model}} &
\multicolumn{5}{c|}{\textbf{GQA}} &
\multicolumn{5}{c|}{\textbf{TextVQA}} &
\multicolumn{5}{c}{\textbf{Avg.}} \\
\cmidrule(lr){2-6}
\cmidrule(lr){7-11}
\cmidrule(lr){12-16}

&
\textbf{\#Tok.} &
\textbf{TFLOPs} &
\textbf{Lat.} &
\textbf{Mem.} &
\textbf{Thpt.} &
\textbf{\#Tok.} &
\textbf{TFLOPs} &
\textbf{Lat.} &
\textbf{Mem.} &
\textbf{Thpt.} &
\textbf{\#Tok.} &
\textbf{TFLOPs} &
\textbf{Lat.} &
\textbf{Mem.} &
\textbf{Thpt.} \\
\midrule

LLaVA V1.5-7B~\cite{liu2024improved}
& 576 & 4.41 & \underline{0.11} & 15.3 & 7.4
& \underline{576} & 4.76 & \underline{0.11} & 16.0 & 5.4
& 576 & 4.55 & \underline{0.11} & 15.7 & 6.5 \\

IMP-3.1B~\cite{shao2025imp}
& 729 & 2.40 & 0.12 & 8.4 & \underline{9.2}
& 729 & \underline{2.48} & 0.12 & 8.4 & \underline{7.1}
& 729 & 2.40 & 0.12 & 8.4 & \underline{8.6} \\

Qwen2-VL-2B~\cite{wang2024qwen2}
& \textbf{352} & \underline{1.50} & 0.13 & 9.3 & 5.5
& 977 & 4.14 & 0.56 & 40.8 & 1.4
& \underline{466} & \underline{2.14} & 0.26 & 30.8 & 4.1 \\

InternVL2-2B~\cite{internvl}
& 1699 & 5.10 & 0.24 & 9.3 & 3.7
& 1668 & 5.09 & 0.24 & 8.8 & 3.2
& 1561 & 4.72 & 0.22 & 9.4 & 3.8 \\

FlashSloth-HD-3.2B~\cite{tong2025flashsloth}
& \underline{414} & 2.71 & \underline{0.11}
& \underline{7.8} & 7.1
& \textbf{414} & 2.83 & \underline{0.11}
& \underline{8.1} & 5.6
& \textbf{414} & 2.59 & \underline{0.11}
& \underline{8.0} & 6.5 \\

\textbf{MoR-MLLM-2.7B (Ours)}
& 576 & \textbf{0.42} & \textbf{0.05}
& \textbf{7.2} & \textbf{16.8}
& \underline{\textbf{576}} & \textbf{0.98}
& \textbf{0.06} & \textbf{7.3} & \textbf{14.5}
& 576 & \textbf{0.61} & \textbf{0.05}
& \textbf{7.2} & \textbf{16.5} \\

\bottomrule
\end{tabular}
}
\end{table*}

\subsection{Benchmarks and Metrics}
\vspace{-0.4em}
We assess the effectiveness of the proposed MoR-MLLM across a comprehensive suite of multimodal and VL benchmarks. Specifically, we evaluate on six multimodal datasets, including MMB~\cite{mmb}, MME~\cite{mme}, MM-Vet~\cite{mmvet}, POPE~\cite{pope}, SEED~\cite{seed}, and MathVista~\cite{mathvista}. In addition, four VL datasets are employed, including SQA~\cite{sqa}, GQA~\cite{gqa}, TextVQA~\cite{textvqa}, and DocVQA~\cite{docvqa}, which emphasize understanding, reading, and reasoning capabilities across different domains. Together, these benchmarks provide a broad evaluation of hallucination resistance, multimodal comprehension, and cross-domain reasoning. All experiments are conducted under the unified \emph{lmms-eval}~\cite{lmms-eval} protocol.

\begin{table*}[!t]
\centering
\tabcolsep=4pt
\caption{Quantitative results of MoR-MLLM and recent lightweight MLLMs on six multimodal and three VL benchmarks. MoR-MLLM achieves a strong balance between efficiency and performance, consistently outperforming or matching larger counterparts across diverse evaluation tasks.}
\label{performance}
\footnotesize
\scalebox{1.05}{
\begin{tabular}{lcc|ccccccccc}
\toprule 
\textbf{Model} & \textbf{Params} &\textbf{Data} & 
\textbf{POPE} & \textbf{MME} & \textbf{MMB} & 
\textbf{MM-Vet} & \textbf{SEED$^I$} & \textbf{MathVista}&
\textbf{GQA}&\textbf{SQA}&\textbf{DocVQA} \\ 
\midrule

LLaVA-1.5~\cite{liu2024improved}&7B&1.2M&85.9&1511&64.3&30.5&58.6&-&62.0&66.8&-\\
LLaVA-NeXT~\cite{llavanext}&8B&1.4M&86.5&1519&67.4&-&70.2&34.6&64.2&70.1&-\\
Eagle-X5~\cite{eagle}&7B&1.2M&88.8&1528&68.4&-&73.9&37.0&64.9&70.0&-\\
DeepSeek-VL~\cite{lu2024deepseek}&7B&\textgreater100M&88.1&-&73.2&41.5&70.4&-&-&-&-\\

\midrule

MobileVLM-V2~\cite{chu2024mobilevlm}&1.7B&3.6M&84.3&1302&57.7&-&-&-&59.3&66.7&-\\
LLaVA-Phi~\cite{llavaphi}&3B&1.2M&85.0&1335&59.8&-&-&-&-&68.4&-\\
MobileVLM-V2~\cite{chu2024mobilevlm}&3B&3.6M&84.7&1441&63.2&-&-&-&61.1&70.0&-\\
DeepSeek-VL~\cite{lu2024deepseek}&1.7B&\textgreater100M&87.6&1532&64.6&34.8&66.7&31.1&-&-&-\\
mini-gemini~\cite{minigemini}&2.2B&2.7M&-&1653&59.8&31.1&-&29.4&-&-&34.2\\
PaliGemma~\cite{team2024gemma}&3B&\textgreater1B&-&1686&71.0&33.1&69.6&28.7&-&-&34.2\\
MiniCPM-V-2~\cite{hu2024minicpmunveilingpotentialsmall}&2.8B&\textgreater500M&87.8&1809&69.6&41.0&67.1&38.7&-&80.7&71.9\\
Imp-v1~\cite{shao2025imp}&3B&1.2M&88.0&1434&66.5&-&-&-&-&70.0&-\\

InternVL1.5~\cite{internvl}&2.2B&\textgreater5M&-&1902&70.9&39.3&69.8&41.1&\underline{61.6}&84.9&85.0\\

InternVL2~\cite{internvl}&2.2B&\textgreater5M&85.2&1877&73.2&\underline{44.6}&71.6&46.3&-&\textbf{94.1}&86.9\\

InternVL2.5-4B~\cite{chen2024expanding}&
4B&
-&
-&
-&
\underline{79.3}&
-&
-&
\textbf{64.1}&
-&
-&
-\\

InternVL2.5-2B~\cite{chen2024expanding}&
2B&
-&
\textbf{90.6}&
2138&
74.7&
-&
-&
51.1&
-&
-&
88.7\\

InternVL3~\cite{zhu2025internvl3}&
2B&
-&
\underline{89.6}&
\textbf{2221}&
\textbf{81.1}&
-&
-&
57.0&
-&
-&
88.3\\

MM1.5~\cite{mm1}&3B&\textgreater45M&88.1&1798&-&41.0&\underline{72.4}& 44.4&-&82.1&87.7\\

Qwen2-VL~\cite{wang2024qwen2}&2B&-&-&1872&74.9&\textbf{49.5}&-&43.0&-&-&\textbf{90.1}\\

Qwen2.5-VL~\cite{bai2025qwen25vltechnicalreport}&
3B&
-&
-&
\underline{2157}&
79.1&
-&
-&
\underline{61.2}&
-&
-&
-\\

FlashSloth~\cite{tong2025flashsloth}&3.2B&3.7M&86.3&1702&73.0&41.9&68.0&42.5&61.1&\underline{88.6}&48.6\\

MoE-LLaVA~\cite{lin2024moellavamixtureexpertslarge}&3.6B&-&86.3&1423&65.2&34.3&-&-&61.4&68.5&-\\

\midrule

\cellcolor{cGrey}MoR-MLLM (ours)&
\cellcolor{cGrey}3.6B&
\cellcolor{cGrey}2.2M&
\cellcolor{cGrey}87.7&
\cellcolor{cGrey}1798&
\cellcolor{cGrey}69.9&
\cellcolor{cGrey}39.9&
\cellcolor{cGrey}\textbf{72.7}&
\cellcolor{cGrey}42.7&
\cellcolor{cGrey}\textbf{62.3}&
\cellcolor{cGrey}70.4&
\cellcolor{cGrey}\underline{89.2}\\

\bottomrule
\end{tabular}}
\end{table*}

\subsection{Quantitative Analysis}
Following FlashSloth~\cite{tong2025flashsloth}, we quantitatively evaluate MoR-MLLM from three perspectives: training efficiency, inference efficiency, and task performance. For the baselines evaluated in FlashSloth~\cite{tong2025flashsloth}, we report the corresponding published results to maintain consistency with the established evaluation setting. Both the reported baseline results and our experiments are obtained on eight NVIDIA RTX 4090 GPUs, ensuring a fair comparison under the same hardware environment.
\subsubsection{Comparative Analysis of Computational Efficiency}
We evaluate the efficiency of MoR-MLLM in comparison to several SoTA MLLMs, using the LLaVA-1.5-7B~\cite{liu2024improved} model as the reference baseline.

\noindent\textbf{Inference Efficiency.}
Table~\ref{inference effeciency} presents a detailed comparison across five representative benchmarks. 
MoR-MLLM achieves a new SoTA trade-off between computational cost and performance. 
It requires significantly fewer floating-point operations and much less GPU memory than all existing MLLMs of similar scale. Although the total number of tokens remains the same, MoR-MLLM provides faster responses, achieving approximately twice the speed of LLaVA~\cite{liu2024improved} and nearly five times that of InternVL2~\cite{internvl} in terms of latency. Throughput also increases by more than fifty percent under comparable GPU memory budgets. These results confirm that the combination of recursive parameter sharing and token-adaptive recursion enables efficient and fine-grained computation.

The efficiency improvements of MoR-MLLM are consistent across
five tasks. Compared with LLaVA-1.5-7B, MoR-MLLM reduces
TFLOPs by 84.6\%, 89.1\%, 89.8\%, 90.5\%, and 79.4\% on
MME, MMBench, POPE, GQA, and TextVQA, respectively. It reduces
GPU memory usage by 52.9\%--57.3\% and first-token latency by
45.5\%--63.6\% across tasks, while increasing throughput by
127.0\%--182.0\%. The FLOPs reductions are most pronounced on GQA, POPE, and MMBench,
whereas TextVQA shows the smallest but substantial reduction of
79.4\%. On average, MoR-MLLM reduces TFLOPs, memory usage, and
latency by 86.6\%, 54.1\%, and 54.5\%, respectively, while improving
throughput by 153.8\%. These results demonstrate that the efficiency
gains are not limited to a specific benchmark, although their magnitude
varies with the task-dependent computation allocated by the router.
\begin{table}[!t]
\caption{Ablation study on balancing coefficient $\lambda_\text{ent}$ and number of experts $m$.}
\centering
\small
\begin{tabular}{@{}cccccccccc@{}}
\toprule
\multirow{2}{*}{\textbf{Hyper-parameters}} & \multirow{2}{*}{\textbf{Value}} & \multirow{2}{*}{\textbf{GQA}} & \multirow{2}{*}{\textbf{POPE}} & \multirow{2}{*}{\textbf{MME}} & \multirow{2}{*}{\textbf{MMB}} \\
& & & & & \\ \midrule
\multirow{3}{*}{$\lambda_\text{ent}$} 
& 0.10    & \textbf{62.3} & \textbf{87.7} & \textbf{1798.0}    & \textbf{69.9} \\
& 0.30        & 62.1         & 87.5          & 1783.3       & 69.7                 \\
& 0.50           & 62.0         & 87.6          & 1779.9       & 69.7                 \\ \midrule
\multirow{3}{*}{number $m$} 
& 3                  & 61.7         & 87.4          & 1740.3       & 69.3                 \\
& 5  & \textbf{62.3} & \textbf{87.7}    & \textbf{1798.0} & \textbf{69.9}           \\
& 10                  & 61.7         & 87.5          & 1741.0       & 69.2                \\
\bottomrule
\end{tabular}
\label{hyper}
\end{table}

\noindent\textbf{Training Efficiency.}
We present a comparative analysis of the training expenditures for
MoR-MLLM and the other four MLLMs. For a fair comparison, all models in Fig.~\ref{training} follow the same software stack, optimization settings, precision mode, hardware configuration, and metric definitions.
As shown in Fig.~\ref{training}, MoR-MLLM achieves a shorter training
timeline than the other MLLMs across both the pre-training and
fine-tuning stages. In total, its pre-training and fine-tuning require
only 65 GPU hours, representing reductions of approximately 48\% and
24\% compared with LLaVA~\cite{liu2024improved} and
IMP~\cite{shao2025imp}, respectively. The dynamic recursion mechanism
allows visual tokens to selectively pass through the recursion blocks,
thereby reducing training computation. MoR-MLLM also requires the
least GPU memory in both stages, using 30 GB during pre-training and
46.8 GB during fine-tuning. MoR-MLLM reduces both training time and
GPU memory consumption without compromising training stability.
\subsubsection{Performance Evaluation Against Existing MLLMs}
We evaluate MoR-MLLM against a range of leading tiny MLLMs across 9 challenging benchmarks covering both multimodal and VL tasks. As summarized in Table~\ref{performance}, MoR-MLLM achieves highly competitive or superior results while requiring substantially less training time and GPU memory. 
To further strengthen the comparison with recent lightweight MLLMs, we include additional baselines, including InternVL2.5-4B, InternVL2.5-2B, InternVL3-2B, and Qwen2.5-VL-3B.
Trained on only 2.2M samples, it attains the best performance on GQA and SEED, and ranks second on DocVQA. Despite being compared with these recent lightweight MLLMs, MoR-MLLM maintains a strong balance between efficiency and performance across diverse multimodal and VL benchmarks.
For multimodal reasoning tasks, it performs strongly on SEED, matching InternVL2 and Qwen2-VL while consistently surpassing MoE-LLaVA. These results demonstrate that the proposed MoR architecture and the three-stage training framework effectively enable near SoTA performance while achieving significant gains in computational efficiency. 

The relative performance gains of MoR-MLLM vary across benchmarks
because different tasks benefit differently from token-adaptive
recursion. GQA requires compositional reasoning over objects,
attributes, and their relations, while SEED covers diverse perception
and multimodal reasoning scenarios. These tasks contain tokens with
heterogeneous computational requirements, allowing the router to
allocate deeper recursion to challenging visual--textual interactions
while avoiding redundant computation for simpler tokens. Consequently,
MoR-MLLM performs particularly strongly on GQA and SEED.
In contrast, performance on benchmarks dominated by hallucination
detection, visual-text recognition, or specialized mathematical
knowledge depends more strongly on the visual encoder, input resolution,
and task-specific supervision, which are not directly modified by the
MoR mechanism. The results therefore suggest that adaptive recursion
provides larger benefits for tasks requiring compositional and
multi-step multimodal reasoning, while its gains are more limited when
performance is primarily constrained by perception or domain-specific
knowledge.
\subsection{Ablation Study}
\subsubsection{Hyperparameter Sensitivity Analysis}
\begin{figure}[!t]
 \includegraphics[width=\linewidth]{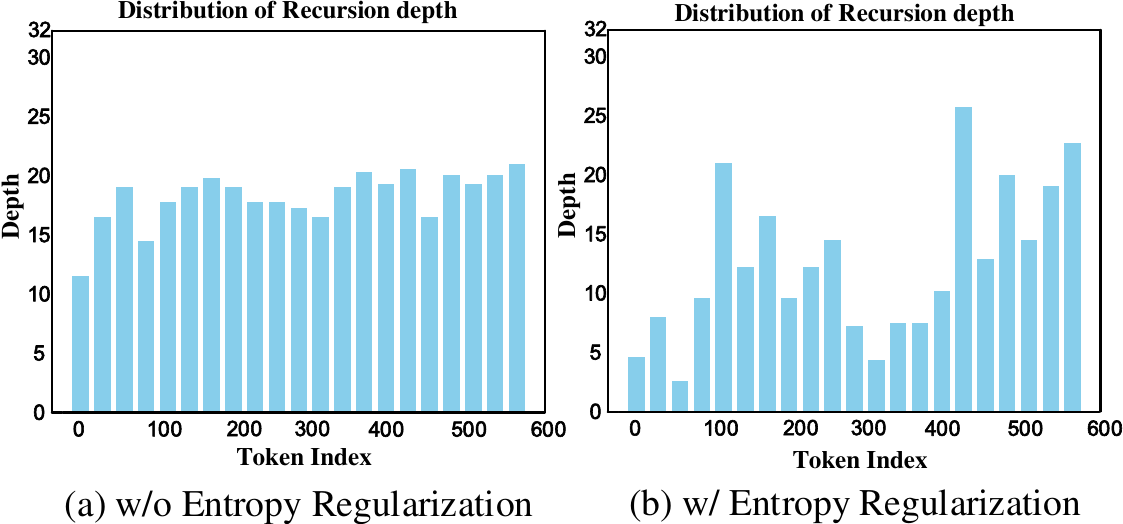}
    \caption{Visualization of recursion-depth distributions with and without entropy regularization.}
    \label{distribution}
\end{figure}

\paragraph{Choice of the Entropy-Regularization Weight $\lambda_\text{ent}$ and Number of Experts $m$}
To further evaluate the proposed architecture, we conduct ablation studies summarized in Table~\ref{hyper}~\ref{tab:mor_tuning_strategy}~\ref{block}. The ablation analysis of the maximum recursion depth $s$ and routing threshold $P_{\beta}$ can be found in Section~\ref{thre}. Table~\ref{hyper} analyzes the effect of the balancing coefficient $\lambda_\text{ent}$. The best performance occurs at $\lambda_\text{ent}=0.10$, which offers a good trade-off between diversity and task-specific optimization. Smaller values weaken regularization, whereas larger ones overemphasize entropy and harm accuracy. As shown in Table~\ref{hyper}, when varying the number of experts, we find that using five achieves the optimal balance between recursion diversity and optimization stability. 

\paragraph{Choice of the Maximum Recursion Depth $S$}
\label{depth}
The maximum recursion depth $S$ defines the available recursion
capacity rather than a fixed number of recursive steps applied to
every token or task. At each recursion step, the router determines
whether each token should continue based on its current hidden
representation. Consequently, different inputs and tasks can produce
different effective recursion-depth distributions within the same
upper bound. We adopt a fixed upper bound to ensure a predictable worst-case
computation budget, while retaining token-level adaptive computation
through the learned router. As shown in Table~\ref{tab:depth_ablation}, increasing $S$ from
8 to 32 improves VQAv2, GQA, and MMBench from 75.2, 59.4, and 63.1
to 77.5, 61.4, and 65.6, respectively. Increasing $S$ further to 48
does not yield additional improvements, with corresponding results of
77.4, 61.2, and 65.5, while GPU memory increases from 7.20 GB to
8.59 GB. Therefore, $S=32$ provides sufficient recursion capacity
for adaptive routing while achieving the best balance between
performance and computational cost.
\label{sec:depth}
\begin{table}[!t]
  \small
  \setlength\tabcolsep{1.2mm}
  \caption{Effect of the maximum recursion depth $S$. Accuracy is reported on VQAv2, GQA, and MMBench, with relative GPU memory normalized to the $S{=}32$ setting.}
  \vskip 0.1in
  \label{tab:depth_ablation}
  \centering
  \begin{tabular}{c|ccc|cc}
\toprule
\textbf{Depth $S$} & \textbf{VQAv2} & \textbf{GQA} & \textbf{MMBench} & \textbf{Mem (GB)}  \\
\midrule
8  & 75.2 & 59.4 & 63.1 & \textbf{5.46}  \\
16 & 76.8 & 60.2 & 64.7 & 6.13  \\
24 & 77.1 & 61.0 & 65.3 & 6.82  \\
32 & \textbf{77.5} & \textbf{61.4} & \textbf{65.6} & 7.20  \\
48 & 77.4 & 61.2 & 65.5 & 8.59 \\
\bottomrule
\end{tabular}
\vskip -0.1in
\end{table}

\paragraph{Ablation on Routing Threshold $P_{\beta}$}
\label{thre}
In the MoR block, the routing threshold $P_{\beta}$ controls the percentile cutoff of routing scores $G^s$ at each recursion step, determining which tokens are forwarded to deeper recursions. 
A smaller $\beta$ value retains fewer tokens (higher sparsity and lower computation), while a larger value increases recursion depth and computational cost. To quantitatively measure routing sparsity, we define the sparsity metric as the ratio of inactive tokens during recursive computation:
\begin{equation}
\mathrm{Sparsity} = 1 - \frac{1}{S T} \sum_{s=1}^{S}\sum_{t=1}^{T} \mathbb{I}\left(g_t^s > P_{\beta}(G^s)\right),
\label{eq:sparsity}
\end{equation}
where $T$ is the total number of tokens, $S$ is the maximum recursion depth, and $\mathbb{I}(\cdot)$ is the indicator function that equals $1$ if the condition holds. 
A higher sparsity value indicates that fewer tokens are activated across recursion steps, leading to reduced computation. We vary $\beta \in \{60, 70, 80, 90\}$ to examine the trade-off between accuracy and efficiency. As summarized in Table~\ref{tab:beta_ablation}, the model achieves the best balance at $\beta = 80$, corresponding to a sparsity of $0.62$, which reduces FLOPs by approximately 87\% while maintaining peak accuracy across all benchmarks.
This choice is consistent across GQA, POPE, MME, and MMBench, with $\beta=80$ achieving the highest performance on all four benchmarks. Lower thresholds further reduce TFLOPs but consistently decrease performance, with larger drops at $\beta=60$, whereas increasing $\beta$ to 90 raises TFLOPs without improving any task. Thus, we observe no alternative threshold that benefits a specific benchmark while degrading others, indicating that $\beta=80$ provides a robust accuracy--efficiency trade-off across datasets.

\begin{table}[!t]
  \small
  \setlength\tabcolsep{1.2mm}
\caption{Ablation on routing percentile threshold $P_{\beta}$. Lower values correspond to higher sparsity (fewer active tokens). The model achieves the best balance between accuracy and efficiency at $\beta=80$, which is adopted as the default setting in all main experiments.}
\label{tab:beta_ablation}
\begin{tabular}{c|c|cccc|c}
\toprule
$P_{\beta}$ (\%) & Sparsity $\downarrow$ & GQA & POPE & MME & MMB & TFLOPs $\downarrow$ \\
\midrule
60 & \textbf{0.42} & 61.5 & 86.9 & 1748.2 & 68.1 & \textbf{0.48} \\
70 & 0.55 & 62.0 & 87.3 & 1770.6 & 69.0 & 0.54 \\
80 & 0.62 & \textbf{62.3} & \textbf{87.7} & \textbf{1798.0} & \textbf{69.9} & 0.61 \\
90 & 0.76 & 62.2 & 87.4 & 1786.5 & 69.3 & 0.72 \\
\bottomrule
\end{tabular}
\end{table}

\subsubsection{Effectiveness of the MoR-Tuning Strategy} 

\begin{table}[!t]
\caption{
Ablation study of the three-stage MoR-Tuning strategy.
ER denotes entropy regularization.
The one-stage baselines use the same 60K-step training budget,
data sequence, and optimization settings as the full pipeline.
}
\label{tab:mor_tuning_strategy}
\centering
\small
\setlength{\tabcolsep}{3.2pt}
\renewcommand{\arraystretch}{1.08}
\resizebox{\columnwidth}{!}{
\begin{tabular}{lcccccc}
\toprule
\textbf{Training Strategy}
&
\textbf{ER}
&
\textbf{Steps}
&
\textbf{GQA}
&
\textbf{POPE}
&
\textbf{MME}
&
\textbf{MMB}
\\
\midrule

After Stage I
&
N/A
&
10K
&
59.3
&
85.7
&
1674.4
&
63.1
\\

After Stages I--II
&
N/A
&
30K
&
59.9
&
86.2
&
1710.8
&
64.3
\\

\midrule

One-stage Joint Training
&
No
&
60K
&
60.9
&
86.9
&
1748.6
&
66.1
\\

One-stage Joint Training
&
Yes
&
60K
&
61.2
&
87.2
&
1763.5
&
66.8
\\

\midrule

MoR-Tuning (Full)
&
No
&
60K
&
62.1
&
87.5
&
1774.3
&
67.7
\\

MoR-Tuning (Full)
&
Yes
&
60K
&
\textbf{62.3}
&
\textbf{87.7}
&
\textbf{1798.0}
&
\textbf{69.9}
\\

\bottomrule
\end{tabular}
}
\arrayrulecolor{black}
\end{table}
\begin{figure}[!t]
 \includegraphics[width=\linewidth]{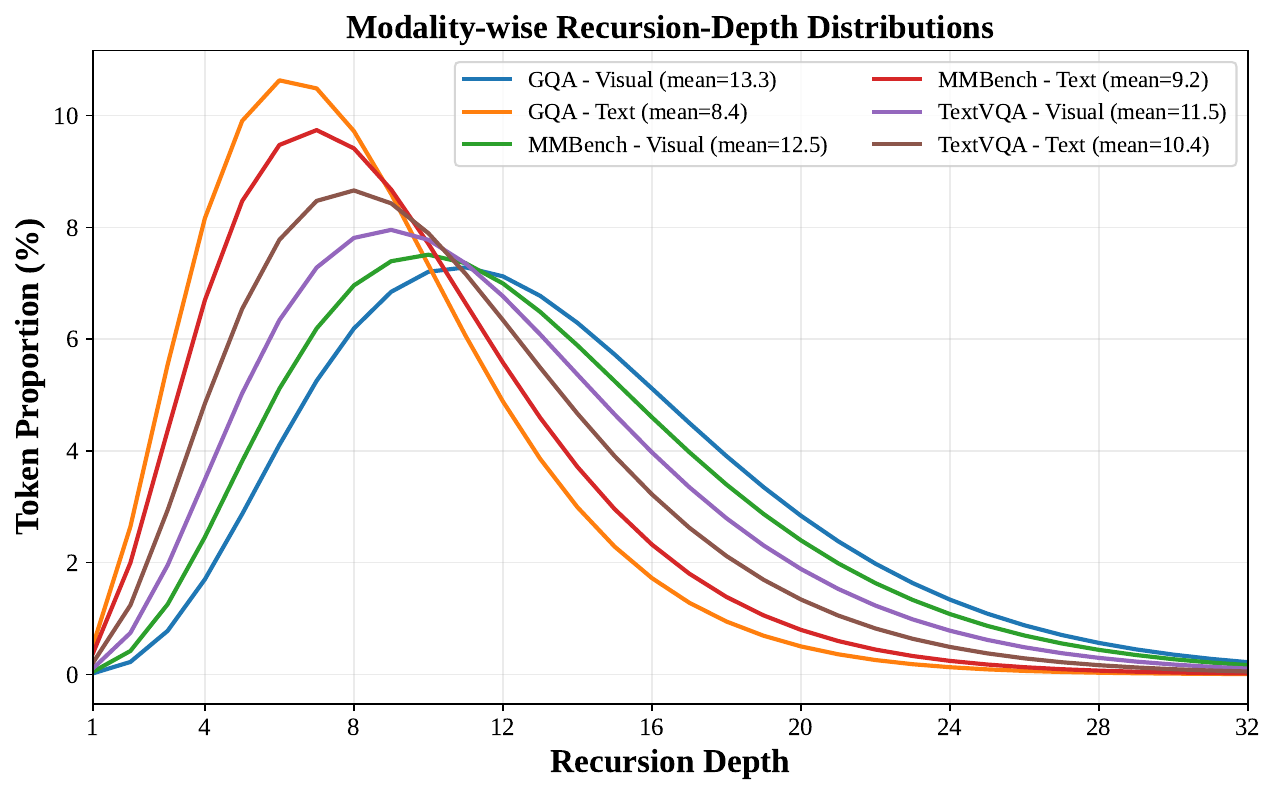}
 \caption{Modality-wise recursion-depth distributions on representative multimodal benchmarks. Visual and textual tokens exhibit distinct depth preferences, while the magnitude of the difference varies across tasks, indicating that the shared router learns modality- and task-dependent computation allocation.}
 \vspace{-3mm}
\label{modality}
\end{figure}
To evaluate the contribution of the three training stages, we first report the intermediate checkpoints obtained after Stage I and after
Stages I--II. As shown in Table~\ref{tab:mor_tuning_strategy}, completing Stage I achieves
59.3 on GQA, 85.7 on POPE, 1674.4 on MME, and 63.1 on MMBench.
After Stage II, the corresponding results increase to 59.9, 86.2,
1710.8, and 64.3. These results indicate that projection warm-up
first establishes visual--language alignment, while the subsequent
adaptation of the shared recursive modules and routers progressively
improves multimodal performance before end-to-end fine-tuning.
To determine whether the final improvement is caused by the staged
design rather than a larger training budget, we additionally introduce
one-stage joint-training baselines. These baselines use the same
backbone initialization, three training datasets and their order,
total number of optimizer steps (60K), batch sizes, and learning-rate
settings as the full MoR-Tuning pipeline. However, the MoR blocks,
routers, visual projector, and other trainable components are jointly
optimized from the beginning, rather than being progressively
introduced through the three stages.
Without entropy regularization, one-stage joint training obtains
60.9 on GQA, 86.9 on POPE, 1748.6 on MME, and 66.1 on MMBench,
whereas the full MoR-Tuning pipeline reaches 62.1, 87.5, 1774.3,
and 67.7, respectively. With entropy regularization, one-stage joint
training achieves 61.2, 87.2, 1763.5, and 66.8, while the complete
MoR-Tuning strategy further improves the results to 62.3, 87.7,
1798.0, and 69.9.
Entropy regularization provides an additional and complementary
benefit by preventing overly concentrated recursion-depth assignments.
As shown in Fig.~\ref{distribution}, the model trained without entropy
regularization exhibits a concentrated routing distribution, whereas
the entropy-regularized variant produces a more diverse and balanced
depth profile. Together with the consistent performance improvements
reported in Table~\ref{tab:mor_tuning_strategy}, these results indicate that entropy regularization
stabilizes the learned routing behavior and improves model
generalization.
Since the compared one-stage and staged variants use the same total
training budget, the consistent improvements demonstrate that the
main gains arise from progressively decoupling visual--language
alignment, recursive routing adaptation, and downstream multimodal
specialization, rather than merely from additional training.

\subsubsection{Effect of MoR Block Initialization}
Table~\ref{block} compares mean initialization~\cite{mean} with
grouped-sharing initialization~\cite{grouped} for the shared
Transformer modules in the MoR blocks. Grouped sharing improves
performance across all four benchmarks, increasing GQA from
61.8 to 62.3, POPE from 87.4 to 87.7, MME from 1744.1 to 1798.0,
and MMBench from 69.5 to 69.9. The improvement is pronounced
on MME, indicating that the initialization strategy has a
meaningful effect on multimodal representation learning. Unlike mean
initialization, which averages multiple pretrained layers into a
shared module, grouped sharing better preserves the distinct
transformations learned by the original Transformer layers. This
provides a more suitable starting point for recursive parameter sharing
and facilitates subsequent joint adaptation of the shared modules and
routers during MoR-Tuning.
\label{ablation}
\begin{table}[!t]
\caption{Ablation study on the MoR block initialization.}
\centering
\small
\begin{tabular}{lcccc}
\toprule
\textbf{MoR Block Initialization} & \textbf{GQA} & \textbf{POPE} & \textbf{MME} & \textbf{MMB} \\
\midrule
Mean Initialization & 61.8 & 87.4 & 1744.1 & 69.5 \\
Grouped Sharing & \textbf{62.3} & \textbf{87.7} & \textbf{1798.0} & \textbf{69.9} \\
\bottomrule
\end{tabular}
\label{block}
\end{table}
\begin{figure}[!t]
 \includegraphics[width=\linewidth]{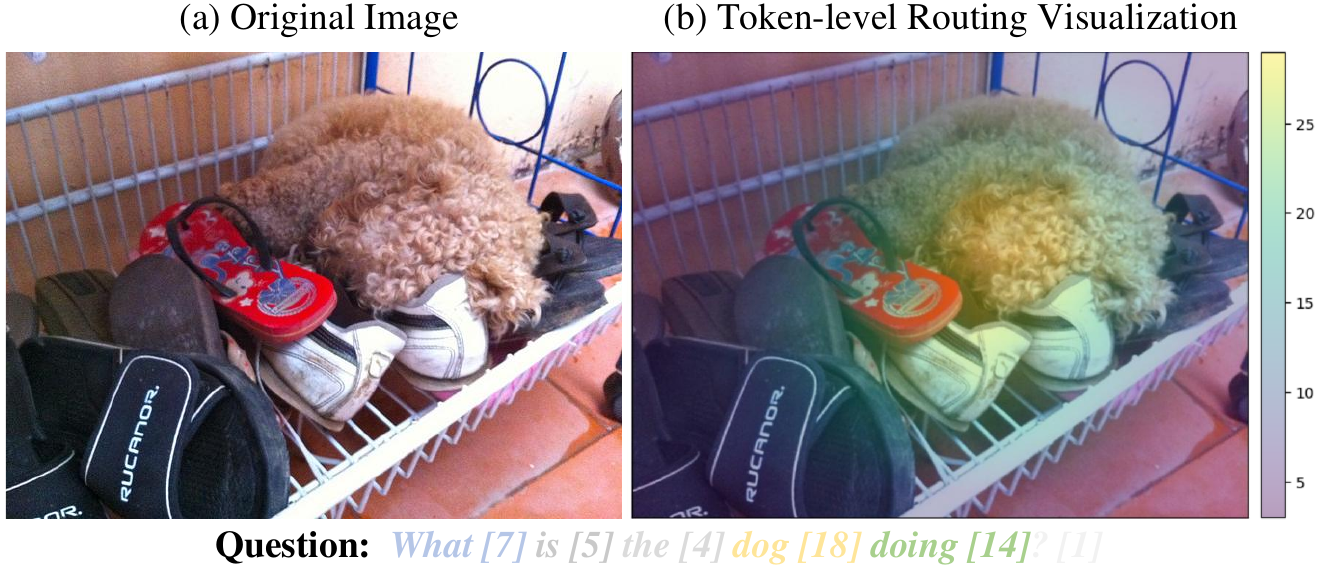}
\caption{
Token-level routing visualization on a representative VQA example.
The heatmap shows the mean recursion depth of visual patches, while the numbers following textual tokens indicate their corresponding recursion depths.
The model allocates deeper computation to task-relevant image regions and content-bearing words, while background regions and function words are processed at shallower depths.}
    \label{token-level}
\end{figure}
\begin{figure*}[!t]
 \centering
 \includegraphics[scale=0.34]{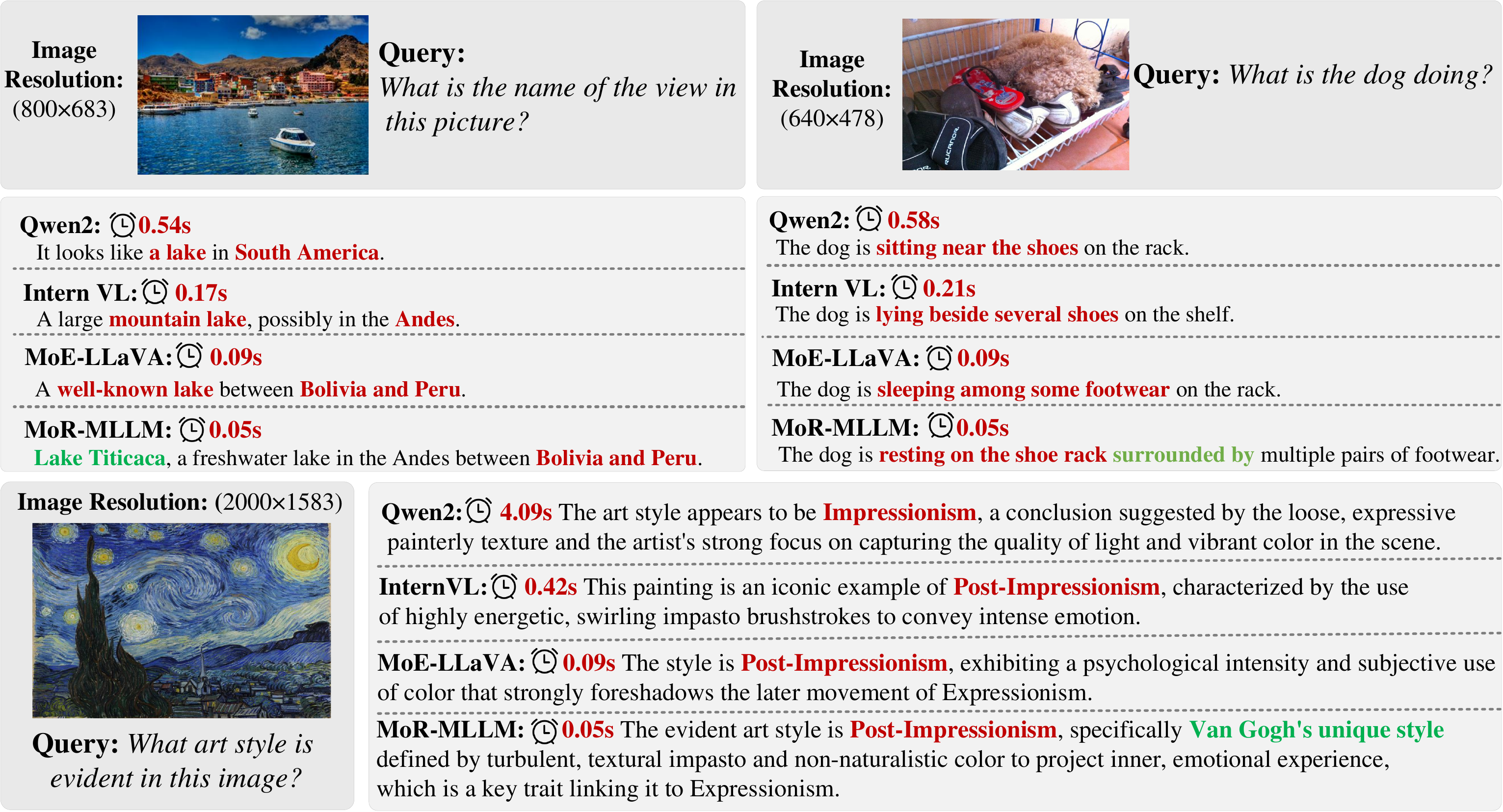}
  \caption{Qualitative comparisons among MoR-MLLM, Qwen2-VL, InternVL2, and MoE-LLaVA on visual question answering and reasoning tasks. MoR-MLLM delivers faster responses with comparable or superior interpretive accuracy, highlighting its advantage in both reasoning quality and computational efficiency.}
\label{visual compare}
\end{figure*}
\subsubsection{Comparison of Routing Strategies}
We compare expert-choice and token-choice routing, together with
visual-only, text-only, and separate modality-specific routing, using
the same backbone, MoR-Tuning pipeline, training data, optimization
schedule, and matched computation budget. As shown in
Table~\ref{tab:routing_strategy}, expert-choice routing achieves
62.3 on GQA, 87.7 on POPE, 1798.0 on MME, and 69.9 on MMBench,
outperforming token-choice routing, which obtains 60.8, 86.9,
1746.5, and 66.7, respectively. Restricting adaptive recursion to
visual or textual tokens leads to lower performance, while
separate modality-specific routers achieve 62.1, 87.6, 1788.7, and
69.5, respectively, remaining slightly below the shared expert-choice
router. Unlike token-choice routing, which determines each token's
complete recursion depth from its initial representation, expert-choice
routing re-evaluates token importance at each recursion step using
progressively updated hidden representations. The shared
router can learn different computation requirements for visual and
textual tokens without separating the routing functions by
modality. This step-wise flexibility and consistently stronger
performance support our adoption of shared expert-choice routing in
MoR-MLLM.

\begin{table}[!t]
\centering
\caption{Comparison of routing strategies under the same training
configuration and approximately matched computation budget.}
\label{tab:routing_strategy}
\resizebox{\columnwidth}{!}{
\begin{tabular}{lccccc}
\toprule
Routing Strategy & GQA & POPE & MME & MMB & TFLOPs $\downarrow$ \\
\midrule
Token-choice
& 60.8 & 86.9 & 1746.5 & 66.7 & 0.61 \\

Visual-only
& 61.7 & 87.3 & 1767.8 & 68.8 & 0.60 \\

Text-only
& 61.1 & 87.0 & 1751.2 & 67.9 & 0.60 \\

Separate Modality Routers
& 62.1 & 87.6 & 1788.7 & 69.5 & 0.62 \\

Shared Expert-choice (Ours)
& \textbf{62.3} & \textbf{87.7} & \textbf{1798.0}
& \textbf{69.9} & 0.61 \\
\bottomrule
\end{tabular}}
\arrayrulecolor{black}
\end{table}





\begin{table}[!t]
\caption{
Controlled ablation of the MoR mechanism.
All variants use the same vision encoder, visual projector,
language backbone, training data, and optimization schedule.
The fixed-depth recursive baseline uses a uniform recursion depth
selected to match the computation budget of MoR-MLLM.
}
\label{tab:mor_attribution}
\centering
\small
\setlength{\tabcolsep}{3.2pt}
\renewcommand{\arraystretch}{1.08}
\resizebox{\columnwidth}{!}{
\begin{tabular}{lccccccc}
\toprule
\textbf{Variant}
&
\textbf{Sharing}
&
\textbf{Routing}
&
\textbf{GQA}
&
\textbf{POPE}
&
\textbf{MME}
&
\textbf{MMB}
&
\textbf{TFLOPs}
\\
\midrule

Dense MLLM
&
No
&
No
&
\textbf{62.5}
&
\textbf{87.8}
&
\textbf{1804.6}
&
\textbf{70.2}
&
2.40
\\

Fixed-depth Recursive
&
Yes
&
No
&
60.9
&
86.8
&
1745.3
&
66.9
&
0.62
\\

MoR-MLLM
&
Yes
&
Yes
&
62.3
&
87.7
&
1798.0
&
69.9
&
\textbf{0.61}
\\

\bottomrule
\end{tabular}
}
\arrayrulecolor{black}
\end{table}
\subsubsection{Controlled Attribution of the MoR Mechanism}
To isolate the contribution of MoR from the other architectural
components, we compare MoR-MLLM with dense and fixed-depth recursive
baselines using the same vision encoder, visual projector, Phi-2
backbone, training data, and optimization schedule. As shown in
Table~\ref{tab:mor_attribution}, the dense baseline achieves slightly
higher performance but requires 2.40 TFLOPs, whereas MoR-MLLM retains
comparable performance with only 0.61 TFLOPs. Under a matched
computation budget, MoR-MLLM consistently outperforms the fixed-depth
recursive baseline across all four benchmarks. These controlled
results show that the performance--efficiency gains primarily arise
from token-adaptive recursion rather than changes to the vision
encoder, projector, or language backbone.

\subsubsection{Generalization Across Backbones and Model Scales}
To examine the generality of MoR-MLLM beyond Phi-2-2.7B, we further
apply the same MoR-Tuning strategy and routing mechanism to
LLaMA-3.1-8B~\cite{grattafiori2024llama}. As shown in Table~\ref{tab:backbone_generalization},
MoR-MLLM reduces inference computation from 6.95 to 1.77 TFLOPs
while maintaining performance close to the dense LLaMA-3.1-8B
counterpart. A similar trend is observed with Phi-2-2.7B, where
TFLOPs decrease from 2.40 to 0.61 with only minor performance
changes. These results demonstrate that the proposed recursive adaptation
can transfer beyond Phi-2 to a different model family and a larger
model scale while preserving a similar performance--efficiency
trade-off.
\begin{table}[!t]
\centering
\caption{Generalization of MoR-MLLM across language backbones and model scales.}
\label{tab:backbone_generalization}
\resizebox{\columnwidth}{!}{
\begin{tabular}{llccccc}
\toprule
Backbone & Variant & GQA & POPE & MME & MMB & TFLOPs $\downarrow$ \\
\midrule
Phi-2 (2.7B)~\cite{phi2}
& Dense MLLM
& 62.5 & 87.8 & 1804.6 & 70.2 & 2.40 \\
Phi-2 (2.7B)~\cite{phi2}
& MoR-MLLM
& 62.3 & 87.7 & 1798.0 & 69.9 & 0.61 \\
\midrule
LLaMA-3.1 (8B)~\cite{grattafiori2024llama}
& Dense MLLM
& 64.7 & 88.5 & 1862.8 & 72.6 & 6.95 \\
LLaMA-3.1 (8B)~\cite{grattafiori2024llama}
& MoR-MLLM
& 64.5 & 88.4 & 1854.6 & 72.3 & 1.77 \\
\bottomrule
\end{tabular}}
\arrayrulecolor{black}
\end{table}

\subsubsection{Routing Behavior Analysis}
To examine whether the shared router applies identical recursion
behavior to visual and textual tokens, we separately analyze their
recursion-depth distributions on GQA, MMBench, and TextVQA. For each
input, tokens are divided according to their original modality, and
the final recursion depth assigned to each token is recorded and
aggregated over the evaluation set. As shown in Fig.~\ref{modality},
visual and textual tokens exhibit clearly different recursion-depth
preferences. On GQA, the average recursion depths of visual and textual
tokens are 13.3 and 8.4, respectively, indicating that substantially
more computation is allocated to visual representations. A similar
tendency is observed on MMBench, with average depths of 12.5 and 9.2.
In contrast, the gap becomes considerably smaller on TextVQA, where
visual and textual tokens reach average depths of 11.5 and 10.4,
respectively. This variation suggests that the required computation is
determined jointly by token modality and task characteristics rather
than by a fixed modality-specific depth.
\begin{table}[!t]
\caption{
Layer-wise average recursion depths of visual and textual tokens
across the five MoR blocks.
}
\label{tab:layerwise_routing}
\centering
\small
\setlength{\tabcolsep}{5pt}
\renewcommand{\arraystretch}{1.08}
\resizebox{\columnwidth}{!}{
\begin{tabular}{lccccc}
\toprule
\textbf{Modality} & \textbf{Block 1} & \textbf{Block 2} & \textbf{Block 3} & \textbf{Block 4} & \textbf{Block 5} \\
\midrule
Visual tokens & 15.0 & 14.0 & 12.7 & 11.3 & 9.9 \\
Text tokens   & 8.2  & 8.6  & 9.1  & 9.7  & 10.5 \\
\bottomrule
\end{tabular}
}
\arrayrulecolor{black}
\end{table}

\begin{table}[!t]
\caption{
Average recursion depth across prediction-confidence quartiles.
Lower-confidence samples receive deeper recursion on average.
}
\label{tab:confidence_routing}
\centering
\small
\setlength{\tabcolsep}{8pt}
\renewcommand{\arraystretch}{1.08}
\resizebox{0.85\columnwidth}{!}{
\begin{tabular}{lcccc}
\toprule
\textbf{Confidence Quartile} & \textbf{Lowest} & \textbf{Q2} & \textbf{Q3} & \textbf{Highest} \\
\midrule
Average Recursion Depth & 13.6 & 12.7 & 11.8 & 10.9 \\
\bottomrule
\end{tabular}
}
\arrayrulecolor{black}
\end{table}
We further analyze how the modality-dependent routing behavior evolves
across the five MoR blocks using MMBench. For each token $t$ at MoR
block $b$, its effective recursion depth $d_{t,b}$ is defined as the
number of recursion steps assigned to the token before it exits the
block. We separately average $d_{t,b}$ over all visual and textual
tokens in the evaluation set. As shown in
Table~\ref{tab:layerwise_routing}, the average recursion depth of
visual tokens decreases from 15.0 in Block~1 to 9.9 in Block~5,
whereas that of textual tokens increases from 8.2 to 10.5. The
modality gap is therefore largest in the earlier blocks and gradually
narrows in later blocks. This pattern suggests that visual
representations receive more recursive refinement in earlier blocks,
while the computation allocated to the two modalities becomes
increasingly similar as multimodal representations are progressively
integrated.
We also examine whether recursion depth is associated with input
difficulty using prediction confidence as a proxy. This analysis is
conducted on MMBench, where prediction confidence is defined as the
normalized probability assigned to the model's selected answer option.
We rank all evaluation samples according to their confidence and divide
them into four equally sized quartiles. For each sample, its average
recursion depth is computed by averaging $d_{t,b}$ over all visual and
textual tokens and all five MoR blocks. As shown in
Table~\ref{tab:confidence_routing}, the average recursion depth
decreases monotonically from 13.6 for the lowest-confidence quartile
to 12.7, 11.8, and 10.9 for increasingly higher-confidence groups.
This inverse relationship indicates that relatively uncertain inputs
tend to receive more recursive computation.

To further reveal which individual tokens receive additional
computation, Fig.~\ref{token-level} provides a token-level routing visualization on a
representative VQA example. The heatmap reports the mean recursion
depth of visual patches, while the numbers associated with textual
tokens indicate their recursion depths. Deeper recursion is
concentrated on task-relevant image regions, whereas background regions
receive shallower processing. Similarly, semantically informative
content words such as ``dog'' and ``doing'' receive larger recursion
depths than common function words. This visualization provides direct
qualitative evidence that the router allocates computation
non-uniformly according to semantic relevance.
Importantly, MoR-MLLM does not explicitly assign different routing
rules to visual and textual tokens. Instead, the shared router makes
step-wise decisions from their evolving hidden representations. The
modality-separated distributions, layer-wise statistics,
confidence-based analysis, and token-level visualization consistently
show that the learned routing behavior varies with modality, task
characteristics, semantic relevance, and input uncertainty. This
observation further distinguishes MoR-MLLM from a direct application
of uniform recursive computation to concatenated visual and textual
tokens.

\subsection{Qualitative Analysis}
Fig.~\ref{visual compare} provides a comparative visualization of the model predictions from Qwen2, Intern VL, MoE-LLaVA, and MoR-MLLM across diverse reasoning examples. First, MoR-MLLM demonstrates highly responsive performance, achieving significantly lower latency than the other models by a wide margin. Moreover, on several types of VQA tasks, MoR-MLLM delivers performance that is superior to the other three models. For instance, in the top-left example, MoR-MLLM accurately recognizes the landmark as Lake Titicaca and further provides fine-grained attributes, describing it as a freshwater lake located in the Andes between Bolivia and Peru, whereas the other models give only vague or incomplete answers. In the top-right example, MoR-MLLM precisely identifies the dog's behavior as resting on the shoe rack surrounded by multiple pairs of footwear, while other models provide less accurate or less detailed descriptions. Besides, a key distinction emerges in the bottom example: while other models fail, only MoR-MLLM identifies the artist as Van Gogh, highlighting its advantage in fine-grained cultural knowledge. Meanwhile, MoR-MLLM consistently achieves the lowest response latency across all examples, demonstrating its superior inference efficiency while maintaining strong multimodal understanding. Overall, these examples demonstrate that MoR-MLLM achieves a favorable balance between computational efficiency and fine-grained multimodal understanding.

\section{Conclusion}
In this paper, we introduce MoR-MLLM, the first computation-sparse MLLM built upon the Mixture-of-Recursions framework. By integrating recursive parameter sharing with per-token dynamic recursion, MoR-MLLM enables adaptive computation allocation according to token complexity. The proposed three-stage MoR-Tuning strategy and entropy-regularized loss stabilize optimization and encourage diverse routing behaviors across tokens. Extensive experiments across a diverse suite of multimodal benchmarks show that the proposed MoR-MLLM substantially reduces computational costs while maintaining competitive performance under comparable training budgets, thereby empirically validating the effectiveness of our design and highlighting recursive sparsity as a promising direction for future~work.
\printbibliography





\end{document}